\documentclass[10pt,final,conference,letterpaper,romanappendices,twocolumn]{IEEEtran}
\IEEEoverridecommandlockouts
\usepackage{xcolor}
\usepackage{epstopdf}
\usepackage{epsfig}
\usepackage{amsthm}
\usepackage{amsmath,amssymb,amsfonts,amstext,amsbsy,amsopn,dsfont}
\usepackage{enumitem}
\usepackage{cases}
\usepackage{sublabel}
\usepackage{bigints}
\usepackage{sublabel}
\usepackage{array}
\usepackage{cuted}
\usepackage{comment}
\usepackage{algorithm}
\usepackage{algpseudocode}
\newtheorem{theorem}{\textbf{Theorem}}

\newcommand{\defn}{\triangleq}
\newcommand{\dif}{\textmd{d}}

\usepackage[utf8]{inputenc}

\algnewcommand{\Inputs}[1]{%
  \State \textbf{inputs:}{~\raggedright #1}
}
\algnewcommand{\Outputs}[1]{%
  \State \textbf{outputs:}{~\raggedright #1}
}
\algnewcommand{\Initialize}[1]{%
  \State \textbf{initialize:}{~\raggedright #1}
}

\begin{document}
	
\title{Modeling and Analysis of Intermittent Federated Learning Over Cellular-Connected UAV Networks\\
\thanks{The work of C.-H. Liu was supported in part by the U.S. National Science Foundation (NSF) under Award CNS-2006453 and in part by Mississippi State University under Grant ORED 253551-060702. The work of L. Wei was supported in part by the NSF under Award CNS-2006612.}
}
	
\author{Chun-Hung Liu${}^{\ddag}$, Di-Chun Liang${}^{\dagger}$, Rung-Hung Gau${}^{\dagger}$, and Lu Wei${}^{*}$\\ Institute of Communications Engineering, National Yang Ming Chiao Tung University, Hsinchu, Taiwan${}^{\dagger}$\\ Department of Electrical and Computer Engineering, Mississippi State University, USA${}^{\ddag}$\\
Department of Computer Science, Texas Tech University, Lubbock TX, USA${}^{*}$
\\e-mail: chliu@ece.msstate.edu$^{\ddag}$; ldc.cm02g@nctu.edu.tw$^{\dagger}$; gaurunghung@nycu.edu.tw$^{\dagger}$; luwei@ttu.edu${}^{*}$}
	
\maketitle

\begin{abstract}
Federated learning (FL) is a promising distributed learning technique particularly suitable for wireless learning scenarios since it can accomplish a learning task without raw data transportation so as to preserve data privacy and lower network resource consumption. However, current works on FL over wireless networks do not profoundly study the fundamental performance of FL over wireless networks that suffers from communication outage due to channel impairment and network interference. To accurately exploit the performance of FL over wireless networks, this paper proposes a novel intermittent FL model over a cellular-connected Unmanned Aerial Vehicle (UAV) network, which characterizes communication outage from UAV (clients) to their server and data heterogeneity among the datasets at UAVs. We propose an analytically tractable framework to derive the uplink outage probability and use it to devise a simulation-based approach so as to evaluate the performance of the proposed intermittent FL model.  Our findings reveal how the intermittent FL model is impacted by uplink communication outage and UAV deployment. Extensive numerical simulations are provided to show the consistency between the simulated and analytical performances of the proposed intermittent FL model.
\end{abstract}

	
\section{Introduction}\label{Sec:Introduction}
In the recent years, we have witnessed that machine learning (ML) techniques have been dramatically advanced and successfully applied to tackle many real-world problems. The remarkable success of ML is mainly attributed to two key factors -- highly powerful computing and extremely efficient data analytics, yet such a remarkable success in ML significantly relies on whether or not there are enough data to support ML algorithms so as to make them work satisfactorily, which becomes a crucial issue in many ML applications. Due to the proliferation of smart mobile devices, collecting data through them becomes much feasible and easier such that a mobile cellular network has gradually been a huge live database abounding with real-time information, which can be utilized by ML to optimize network operations and managements. Proper and efficient utilization of ML techniques based on data distributed over a massive mobile network becomes an important issue. This is especially true when transporting raw data from all mobile devices to a server in a massive network because it causes many issues, such as network congestion, energy consumption, privacy, security, etc. To avoid transporting a huge amount of distributed data to a server for conducting centralized ML, a distributed learning methodology without raw data transportation, such as federated learning (FL)~\cite{WXHYWCZCXC19,QYYLYCYK19,NSDHSRJH20}, becomes a viable solution.
	
A number of the existing FL algorithms were developed with uniformly compressible data and shown to achieve convergence based on the assumption of error-free and reliable data communications between a server and clients. For example, reference \cite{FSSWKRMWS20} proposed a compression protocol that inherits the compression techniques of top-$k$ sparsification and quantization for uplink and downlink communication in FL. In \cite{WSTTSLKKM19}, FL-based multi-access edge computing was studied with limited network resources and it adopted a gradient descent approach to find the optimal trade-off between local update at clients and global aggregation at a server. There are also a number of works in the recent years studying the problem of FL over wireless communication, where many of them approached the problem from the perspective of signal processing. The authors of reference \cite{AMMGD20}, for instance, devised a compressive sensing approach for FL over single-antenna communication systems. The authors of reference~\cite{JYSAMMLJPHV21} proposed a compressive sensing approach for FL over a MIMO communication system, where the server recursively finds the linear minimum-mean-square-error estimate of the transmitted signal by exploiting the sparsity of the signal.  In \cite{ZGSYGSHK21}, the authors studied the over-the-air computation (AirComp, proposed in \cite{ZGWYHK20}) problem with one-bit broadband digital aggregation. Furthermore, very few works studied FL over UAV networks, such as~\cite{PQVZMRRHTW21}. 

\begin{figure*}[!t]
	\centering
	\includegraphics[height=1.6in,width=0.75\linewidth]{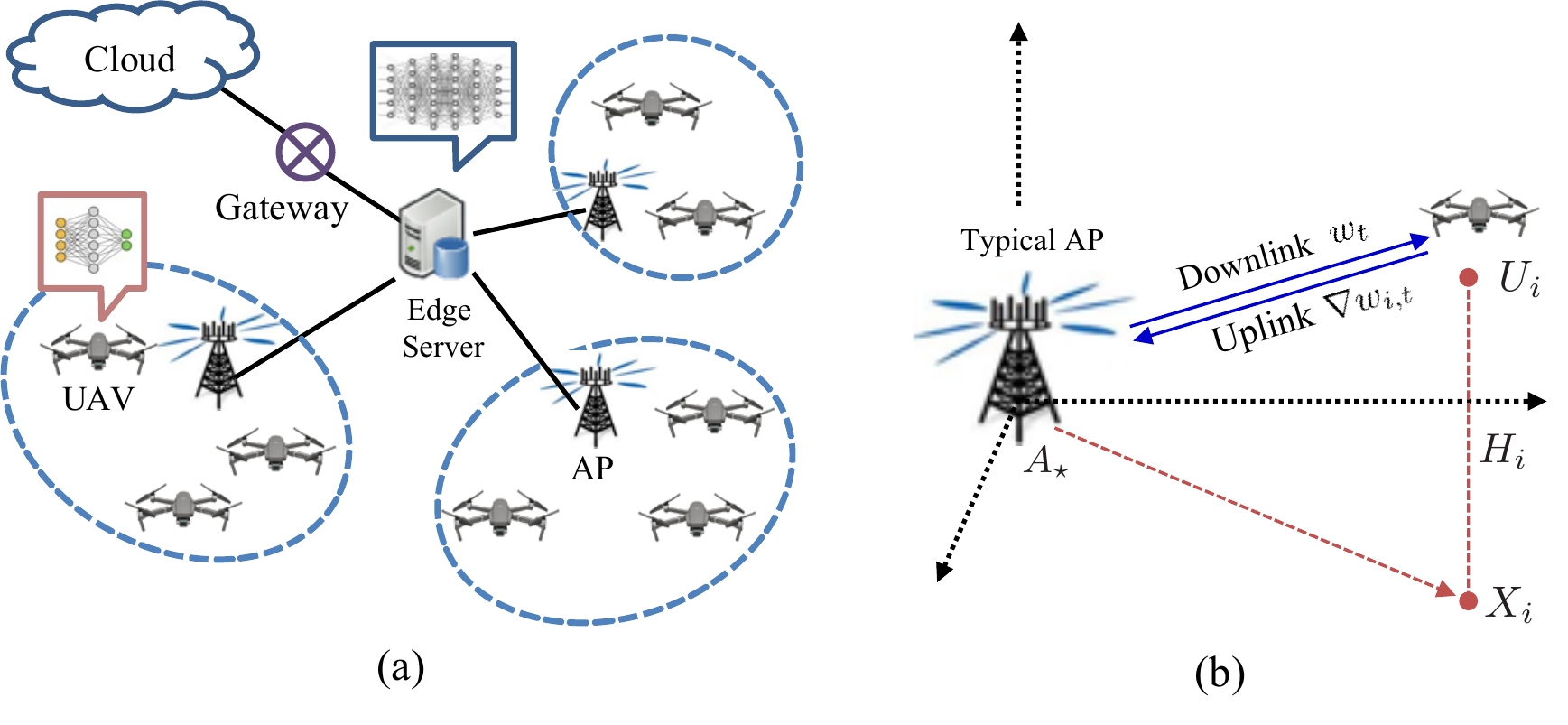}
	\vspace{-0.15in}
	\caption{(a) A cellular-connected UAV network consisting of UAVs (clients), APs, an edge server, and a cloud. Each UAV associates with an AP in order to jointly conduct the proposed intermittent FL with the edge server. (b) A schematic diagram to illustrate a UAV association scenario in which UAV $U_i$ associates with the typical AP located at the origin, denoted by $A_{\star}$, which sends the global model vector $w_t$ to $U_i$ and receives $\nabla w_{i,t}$ from $U_i$.}
	\label{Fig:SystemModel}
	\vspace{-0.15in}
\end{figure*}
	
In these prior works, a fundamental issue of FL over wireless network is far from being fully resolved, that is, data communication between clients and a server may fail due to unreliable wireless transmissions, which leads to communication outage and degrades the convergence performance of FL accordingly. Another crucial issue that was not addressed much in the prior works is the heterogeneity of datasets among different mobile clients. Namely, most of the existing works focus on developing wireless FL algorithms by assuming that all mobile clients possess independent and identically distributed (i.i.d.) datasets. To tackle these two issues, we first propose a more realistic FL model over a cellular-connected UAV network that characterizes uplink communication outage from UAV (clients) to an edge server, which is our first contribution. Our second contribution is to propose a 3D random deployment model of UAVs and use it to develop a tractable framework of analyzing the uplink outage probability of a UAV in the network. Our third contribution is to analytically show that the performance of the proposed intermittent FL over a UAV network can be significantly degraded by communication outage from UAVs to the edge server due to the improper deployments of access points (APs) and UAVs in the network. In addition, intensive numerical simulations are conducted to validate our analytical findings.

\section{System Model}
	
\subsection{Model of a Cellular-Connected UAV Network}\label{SubSec:NetworkModel}
In this paper, we consider a cellular network consisting of an edge server, a tier of APs, and a tier of flying users, i.e., UAVs, which collect data for learning. The edge server is connected to a cloud through a gateway and it  sends data to the cloud whenever it needs the cloud to do large-scale data processing and learning. An illustration of the cellular-connected UAV network is shown in Fig.~\ref{Fig:SystemModel}(a). All the UAVs in the network are assumed to be distributed according to the following 3D point process\footnote{This 3D point process is a generalization of the 3D point process proposed in our previous work~\cite{CHLDCLRHG21} by considering a general distribution of the altitude of each UAV.}:
\begin{align}
\Phi_u \defn \{U_i\in\mathbb{R}^2\times\mathbb{R}_+:U_i=(X_i,H_i),i\in\mathbb{N}_+ \},
\end{align}
where $U_i$, denotes UAV $i$ and its 3D location, $X_i\in\mathbb{R}^2$ is the projection of $U_i$ on the ground, and $H_i\in\mathbb{R}_+$ is the (random) altitude of $U_i$. All the $H_i$'s are i.i.d. and independent of all the $X_i$'s. Fig.~\ref{Fig:SystemModel}(b) shows a typical AP located at the origin and a UAV $U_i$ associating with it. The set of the projections of all the UAVs, i.e., $\{X_i\}$, are assumed to form a 2D  independent Homogeneous Poisson Point Process (HPPP) of density $\lambda_u$, whereas all the APs also form a 2D independent HPPP of $\lambda_a$, which can be expressed as
\begin{align}
\Phi_a \defn \{A_j\in\mathbb{R}^2:j\in\mathbb{N}_+ \},
\end{align}
where $A_j$ denotes AP $j$ and its location. 

Due to the 3D position of a UAV, a wireless link between a UAV and a ground AP can be line-of-sight (LoS) or non-LoS (NLoS). A wireless LoS link between two spatial points means that the link is not visually blocked from one point to the other. For a low-altitude platform of UAV communications, the LoS probability of a wireless link between a UAV and a ground point was proposed in~\cite{AHAKSLS14}. We adopt it in this paper and express it by using the coordinate system in Fig.~\ref{Fig:SystemModel}(b) where a typical AP is located at the origin (denoted by $A_{\star}$) and a UAV is located at $U_i$ as follows:
\begin{align} \label{Eqn:LoSProb}
	\rho(\Theta_{i\star}) \defn \frac{1}{1+c_2\exp(-c_1\Theta_{i\star})},
\end{align}
where $\Theta_{i\star}\defn\tan^{-1}(H_i/\|X_i\|)$ is the elevation angle from the typical AP to UAV $U_i$, $\|X_i\|$ denotes the distance between $X_i$ and the typical AP, $c_1$ and $c_2$ are the environment-related positive coefficients (for rural, urban, etc.). 

Each UAV associates with an AP in the network that provides it with the strongest signal power on average. For example, if UAV $U_i$ in Fig.~\ref{Fig:SystemModel}(b) associates with the typical AP located at the origin, i.e., $A_{\star}$, which satisfies the following expression:
\begin{align} \label{Eqn:AssoScheme}
	A_{\star} &=\arg\max_{j:A_j\in\Phi_a} L_{ij}\|U_i-A_j\|^{-\alpha} \nonumber\\
	&=\arg\max_{j:A_j\in\Phi_a} L_{ij}\left(\|X_i-A_j\|^2+\|H_i\|^2\right)^{-\frac{\alpha}{2}}\nonumber\\
	&\stackrel{(*)}{=} \arg\min_{j:A_j\in\Phi_a} L^{-\frac{1}{\alpha}}_{ij}\|X_i-A_j\|\equiv \mathbf{0},
\end{align}
where $\alpha >2$ denotes the path loss exponent, $L_{ij}\in\{1,\ell\}$ is the LoS link gain between $U_i$ and $A_j$, and $\stackrel{(*)}{=}$ is due to the fact that $\|H_i\|$ does not affect the result of associating with an AP. $L_{ij}$ is a Bernoulli random variable that is one if the link between $U_i$ and $A_j$ is LoS and $\ell\in(0,1)$ otherwise. Note that the transmit power of the APs and the fading effect in each wireless link are not considered in~\eqref{Eqn:AssoScheme} because all the APs are assumed to have the same transmit power and the fading effect in each wireless link is averaged out on the receiver side. Moreover, we assume a densely distributed scenario of UAVs (i.e., $\lambda_u\gg\lambda_a$) in the network such that each AP is almost surely associated with at least one UAV, and thereby each AP is able to deliver the signals  between its UAVs (clients) and the edge server for federated learning. A more realistic model of federated learning between the edge server and the UAVs will be proposed in the following subsection.

\subsection{ Model of Intermittent Federated Learning}\label{SubSec:FedLearnPreSim}
In the UAV network proposed in Section~\ref{SubSec:NetworkModel}, we aim to study the \textit{realistic} performance of FL over the UAV network that undergoes \textit{communication outages} from a UAV  to its (associating) AP, which lead to \textit{non-continuous} FL between the edge server and the UAVs. Thus, we propose an intermittent FL model over the network in Fig.~\ref{Fig:SystemModel}(a) as follows. Suppose there are $K$ UAVs distributed on average in the network and they would like to jointly learn a global model with the edge server in $T$ training rounds. To simply characterize the impact of communication outages on FL over the network, we assume FL is much severely impacted by the communication outages in the uplink direction, i.e., from a UAV to its AP. Such an assumption is reasonable since the communication outages in the downlink direction can be significantly mitigated by multiple broadcasting attempts from an AP to its UAVs during the training process of FL. As such, we propose the model of intermittent FL over the cellular-connected UAV network, as detailed in Algorithm~\ref{Alg:FedUAV}. 

In Algorithm~\ref{Alg:FedUAV}, each UAV $U_i$ is assumed to possess a dataset $\mathcal{D}_i$  given by
\begin{align}
\mathcal{D}_i \defn \{D_{ij}=(x_{ij},y_{ij}), j\in\mathbb{N}_+\},\,\, i\in\{1,2,\ldots, K\},
\end{align}
where $D_{ij}$ denotes data point $j$ in dataset $\mathcal{D}_i$, $x_{ij}$ is the input data vector with an appropriate dimension, $y_{ij}$ is the labeled scalar output corresponding to $x_{ij}$. We assume that all the datasets $\mathcal{D}_i$'s are non-i.i.d and all the UAVs are able to synchronously update their local learning model. At the $t$th round of training, UAV $U_i$ updates its local model vector $w_{i,t}$ according to the following algorithm:
\begin{align}
		\begin{cases}
			w_{i,t} \leftarrow \text{download}_{i\leftarrow S}(w_{t-1})\\
			\nabla w_{i,t} \leftarrow \sum_{j:D_{ij}\in\mathcal{D}_i} \text{SGD}(D_{ij},w_{i,t}) - w_{i,t}~~~
		\end{cases},
\end{align}
where $\text{SGD}(D_{ij},w_{i,t})$ stands for the calculation of Stochastic Gradient Descent with data point $D_{ij}$ and model vector $w_{i,t}$. After receiving the global model vector $w_{t-1}$ from the edge server broadcasted by the AP, UAV $U_i$ updates its local model vector $w_{i,t}$ by SGD so as to find the gradient $\nabla w_{i,t}$. Afterwards, UAV $U_i$ uploads $\nabla w_{i,t}$ to the edge server through the typical AP. To characterize the communication outage from $U_i$ to the typical AP, a Bernoulli random variable $\beta_{i,t}\in\{0,1\}$ is used to model the uplink communication outage from UAV $U_i$ to the typical AP at the $t$th round\footnote{Note that the distribution of $\beta_{i,t}$ is affected by the density of the UAVs transmitting at the same time, which will be elaborated in Section~\ref{Sec:AnalysisUAV}.}. As a result, the data aggregation algorithm at the edge server can be written as
\begin{align}
		\begin{cases}
			\nabla w_{t} \leftarrow \frac{1}{\sum_{i}\beta_{i,t}|\mathcal{D}_i|} \sum_{i} \beta_{i,t}|\mathcal{D}_i|\nabla w_{i,t} \\
			w_{t} \leftarrow w_{t-1} + \nabla w_{t}
		\end{cases}.
\end{align}
Namely, the edge server proportionally combines the received gradients into $\nabla w_t$ based on the sizes of the datasets at the UAVs~\cite{QYYLYCYK19}. As previously pointed out, some local gradients may not be successfully uploaded to the edge server because of uplink communication outage, and thereby the edge server may not be able to aggregate all the local gradients transmitted by all the UAVs. The global model vector $w_t$ at the $t$th round is updated and then broadcast to all the UAVs in the next round. Such an intermittent FL process between the edge server and the $K$ UAVs proceeds until the predesignated number $T$ of training rounds is reached.

To illustrate how the proposed intermittent FL in Algorithm~\ref{Alg:FedUAV} is impacted by uplink communication outage, we will first provide the analyses of the outage probability of the uplink communications from a UAV to its AP in the following section. Next, some simulation results regarding intermittent FL will be provided in Section~\ref{Sec:PreSim} to numerically demonstrate how the performance of intermittent FL is impacted by the uplink communication outages.

\begin{algorithm}[!t]
	\caption{Intermittent Federated Learning}
	\label{Alg:FedUAV}
	\begin{algorithmic}[1]
		\Inputs{initial model vector $w_o$}
		\Outputs{improved model vector $w$}
		\Initialize{the global model is initialized as $w_{0}\leftarrow w_o$.} Each client holds non-i.i.d. dataset $\mathcal{D}_i$ with equal size. $\nabla w_0,\nabla w_{i,0}\leftarrow 0$.
		\For{$t=1,\cdots,T$}
		\For{$i\in\{1,\cdots,K\}$} \textit{\textbf{in parallel}}\footnotemark 
		\State client $i$ does:
		\State $\cdot~ w_{i,t}\leftarrow \text{download}_{i\leftarrow S}(w_{t-1})$
		\State $\cdot~ \nabla w_{i,t} \leftarrow \sum_{j:D_{ij}\in\mathcal{D}_i} \text{SGD}(D_{ij},w_{i,t}) - w_{i,t}$
		\State $\cdot~\text{upload}_{S\leftarrow i}(\nabla w_{i,t})$
		\EndFor
		\State edge server does:
		\State $\cdot~ \nabla w_{t} \leftarrow \frac{1}{\sum_{i}\beta_{i,t}|\mathcal{D}_i|} \sum_{i} \beta_{i,t}|\mathcal{D}_i|\nabla w_{i,t}$
		\State $\cdot~ w_{t} \leftarrow w_{t-1} + \nabla w_{t}$
		\State $\cdot~ w \leftarrow w_{t}$
		\State $\cdot~\text{broadcast}_{i\leftarrow S}(w_t)$
		\EndFor
		\State\Return $w$
	\end{algorithmic}
%	\vspace{-0.2in}
\end{algorithm}
	
\section{Analysis of the Uplink Outage Probability}\label{Sec:AnalysisUAV}
	
%	To demonstrate how the uplink transmission outage is affected by UAV and AP deployments and channel fading, in this section, we propose a UAV deployment model in which tractable analyses of the uplink outage probability can be conducted.

Suppose the network is interference-limited and consider the uplink communication scenario from $U_i$ to the typical AP $A_{\star}$ shown in Fig.~\ref{Fig:SystemModel}(b). As such, the Signal-to-Interference Ratio (SIR) at $A_{\star}$ can be defined as
\begin{align}
\gamma_{i\star} \defn \frac{G_{i\star} L_{i\star}\|A_\star\|^{-\alpha}}{I_{\star}},
\end{align}
where $G_{i\star}$ is the fading channel gain from $U_i$ to $A_{\star}$. Accordingly, the uplink outage probability of a UAV is defined as
\begin{align}\label{Eqn:DefnOutProb}
p_{out}\defn \mathbb{P}[\gamma_{i\star}\leq\eta]=\mathbb{P}[\beta_{i,t}=0],\,\,\text{for all $i$ and $t$},
\end{align}
where $\eta>0$ is the SIR threshold for successful decoding. In this section, we focus on the analysis of $p_{out}$ that affects the performance of the intermittent FL model proposed in Section~\ref{SubSec:FedLearnPreSim}. To facilitate the derivation of $p_{out}$, we first need to introduce two related theorems. The first theorem stated in the following is about the distribution of the path loss of a wireless link from a UAV to its AP.
	
\begin{theorem} \label{Thm:CDFAssoPower}
Suppose UAV $U_i$ associates with the typical AP located at the origin. If $R_{i\star}(r) = \mathbb{P}[L_{i\star}\|U_i\|^{-\alpha} \leq r]$, then for UAV $U_i$ with a given altitude $H_i$ it can be found as given by
\begin{align}
R_{i\star}(r) =\exp\left[-\pi\lambda_a\Upsilon_i(r)\right],
\end{align}
where $\Upsilon_i(\cdot)$ is defined as
\begin{align} \label{Eqn:CDFAssoPower_inner}
\Upsilon_i(r) \defn&
\int_{((\frac{\ell}{r})^{-\frac{2}{\alpha}}-H_i^2)^+}^{(r^{-\frac{2}{\alpha}}-H_i^2)^+}\rho\left(\vartheta_i(y)\right)\mathrm{d}y +\left(\left(\frac{\ell}{r}\right)^{-\frac{2}{\alpha}}-H_i^2\right)^+
\end{align}
in which $(x)^+ \defn \max\{0,x\}$ and $\vartheta_i(y)\defn \tan^{-1}(H_i/\sqrt{y})$.
\end{theorem}
\begin{IEEEproof}
See Appendix~\ref{App:ProofAssoPower}.
\end{IEEEproof}
\noindent Note that $\Upsilon_i(r)$ is a decreasing function of $r$ when the upper limit of the integral in~\eqref{Eqn:CDFAssoPower_inner} is not zero. From the expression in~\eqref{Eqn:CDFAssoPower_inner}, we thus know how $R_{i\star}(r)$ varies with $H_i$, $\lambda_a$, and $\ell$.

Next, we need to analyze the interference received by the typical AP. To make the analysis tractable, we assume all the UAVs associating with the same AP do not use the same resource blocks in the uplink so that the UAVs using the same resource blocks in the network also form a 3D point process. Let $I_{\star}$ be the interference received by the typical AP and it can be defined as
\begin{align}\label{Eqn:DefUplinkInter}
I_{\star} \defn \sum_{k:U_k\in\widetilde{\Phi}_u\backslash U_i} G_{k\star}L_{k\star}\|U_k\|^{-\alpha},
\end{align}
where $\widetilde{\Phi}_u$ is the set of all the UAVs using the same resource block as $U_i$ and $G_{k\star}\sim\exp(1)$ that denotes an exponential random variable with unit mean is the fading channel gain from UAV $U_k\in\widetilde{\Phi}_u$ to the typical AP and and independent of any other random variables in $I_{\star}$ for all $k\in\mathbb{N}_+$. Hence, all $G_{k\star}$'s are i.i.d. In the following theorem, we specifically show the Laplace transform of $I_{\star}$, which is defined as $\mathcal{L}_{I_{\star}}(s)\defn\mathbb{E}[\exp(-sI_{\star})]$ for $s>0$.
\begin{theorem}\label{Thm:LapTranUplinkInter}
The Laplace transform of $I_{\star}$ can be found as
\begin{align} \label{Eqn:LapTranUplinkInter}
\mathcal{L}_{I_{\star}}(s) =\exp\left(-\pi \lambda_a\int^\infty_0\mathcal{I}_G\left(sy^{-\frac{\alpha}{2}},\vartheta(y)\right)\mathrm{d}y\right),
\end{align}
where $\vartheta(y)\defn \tan^{-1}(\frac{H_k}{\sqrt{y}})$ and $\mathcal{I}_G(u,w)$ for $u,w>0$ is defined as
\begin{align}\label{Eqn:LapTranUplinkInter_innerTerm}
\mathcal{I}_G(u,w) \defn& \rho(w)[1-\mathbb{E}\{\mathcal{L}_G(u\cos^\alpha(w))\}]\nonumber\\
&+ (1-\rho(w))[1-\mathbb{E}\{\mathcal{L}_G(\ell u\cos^\alpha(w))\}]
\end{align}
in which $\mathcal{L}_{G}(s)\defn \mathbb{E}[\exp(-s G_{k\star})]$. 
\end{theorem}
\begin{IEEEproof}
See Appendix~\ref{App:ProofLapTranUplinkInter}.
\end{IEEEproof}
\noindent From Theorem~\ref{Thm:LapTranUplinkInter}, we are able to learn how the statistical properties of the uplink interference is affected by the deployments of UAVs and APs. Note that \eqref{Eqn:LapTranUplinkInter} is a function of $\lambda_a$ since the density of the UAVs using the same uplink resource blocks is also the same as that of the APs. 

Using the results in Theorems~\ref{Thm:CDFAssoPower} and~\ref{Thm:LapTranUplinkInter}, we can derive the uplink outage probability as shown in the following theorem.
\begin{theorem}\label{Thm:UplinkOutProb}
According to the uplink outage probability defined in~\eqref{Eqn:DefnOutProb}, it can be explicitly found as
\begin{align} \label{Eqn:UplinkOutProb}
p_{out} =1 - \mathbb{E}\left\{\exp\left(-\pi \lambda_a\int^\infty_0\mathcal{I}_G\left(\frac{\eta y^{-\frac{\alpha}{2}}}{R'_{i\star}},\vartheta_i(y)\right)\mathrm{d}y\right)\right\},
\end{align}
where $R'_{i\star} (r) \defn \frac{\mathrm{d}R_{i\star}(r)}{\mathrm{d}r}=-\pi \lambda_a\Upsilon_i'(r) \exp\left(-\pi\lambda_a\Upsilon_i(r)\right)$.
\end{theorem}
\begin{IEEEproof}
		%See Appendix \ref{App:ProofUplinkOutProb}
According to $p_{out}$ in~\eqref{Eqn:DefnOutProb}, it can be rewritten as
\begin{align*}
p_{out} &=\mathbb{P}\left[G_{i\star}\leq\frac{\eta I_{\star}}{L_{i\star}\|A_\star\|^{-\alpha}}\right]
			= 1 - \mathcal{L}_{I_{\star}}\left(\frac{\eta}{L_{i\star}\|A_\star\|^{-\alpha}}\right).
\end{align*}
By employing the results in Theorems~\ref{Thm:CDFAssoPower} and~\ref{Thm:LapTranUplinkInter} to the above expression of $p_{out}$, the explicit result of $\mathcal{L}_{I_{\star}}(\cdot)$ can be found so that $p_{out}$ in~\eqref{Eqn:UplinkOutProb} is readily obtained.
\end{IEEEproof}
\noindent Theorem~\ref{Thm:UplinkOutProb} clearly indicates how $p_{out}$ relates to $\lambda_a$ and $\ell$, which reveals the fact that the distribution of $\beta_{i,t}$ is affected by $\lambda_a$ and $\ell$ as well. In other words, the realistic performance of FL over a UAV network depends on how densely UAVs are distributed in the network and how they are positioned in the sky. In the following section, some numerical results will be provided to demonstrate this observation. 
	
\section{Numerical Results and Discussions}
\begin{table}[!t]
		\centering
		\caption{Network Parameters for Simulation}\label{Tab:SimPara}
		\begin{tabular}{|c|c|}
			\hline Parameter& Value\\ 
			\hline UAV Density $\lambda_u$ (UAVs/m$^2$) & $1\times 10^{-5}$ \\ 
			\hline AP Density $\lambda_a$ (APs/m$^2$) & $\lambda_u/[10, 300]$ \\
			\hline UAV Height $H_i$ (m) & $100$ \\ 
			\hline SIR Threshold $\eta$ & $0.5$ \\
			\hline Path-loss Exponent $\alpha$ & $2.75$ \\
			\hline $(c_1, c_2)$ in \eqref{Eqn:LoSProb} for urban & $(0.1581, 43.9142)$ \\
			\hline Attenuation Gain of NLoS Channels $\ell$ & $0.25$ \\
			\hline Size of Training Dataset at each UAV $|D_i|$ & $20$\\
			\hline Average Number of UAVs (clients) $K$ & $10, 30, 50$\\
			\hline Number of Training Rounds $T$ & $200$\\
			\hline
		\end{tabular} 
%	\vspace{-0.2in}
\end{table}
	
This section provides some simulation results of the proposed intermittent FL by considering the MNIST datasets of handwritten digits stored at the UAVs. We numerically evaluate the performance of the proposed intermittent FL by using the metric of \emph{learning accuracy}, which is defined as the rate of using the global model $w$ learned by the proposed intermittent FL to successfully classify the images of handwritten digits in the entire dataset stored in the network\footnote{Thus, the entire dataset in the network is the union of all the local datasets stored at the UAVs.}. We first present and discuss the numerical results of the learning accuracy of the proposed intermittent FL over the network and afterwards we show the numerical results of how the learning accuracy is influenced by the deployment densities of the APs and UAVs. The values of the network parameters used for simulation are listed in Table~\ref{Tab:SimPara}.
	
\subsection{Numerical Results of the Proposed FL Model} \label{Sec:PreSim}
	
In this subsection, we provide numerical results regarding how the performance of the proposed intermittent FL model is influenced by uplink communication outages. All the uplink channels are assumed to experience independent block fading such that all $\beta_{i,t}$'s are i.i.d. for all $i$ and $t$. Also, the datasets at the UAVs are assumed to be of the same quality and size. To make the learning processes at different UAVs consistent, all the UAVs adopt the same architecture of a deep neural network to perform local learning, that is, the neural networks at different UAVs adopt the same batch sizes, the same number of the neurons in each layer, and the same number of hidden layers. The learning cases of i.i.d. and non-i.i.d. datasets are both considered in the simulation. Each UAV collects a dataset where the ratio of training data size to testing data size is 2:1. To make each learning case fairly compared, all the ten image classes are uniformly collected by the UAVs. For example, if there are $50$ UAVs involved in the training processing of FL, the image class of handwritten digit ``$3$" is collected by exactly $5$ UAVs among the $50$ UAVs, which happens to the other nine image classes likewise.
	
\begin{figure}[!t]
		\includegraphics[width=1.0\linewidth,height=1.75in]{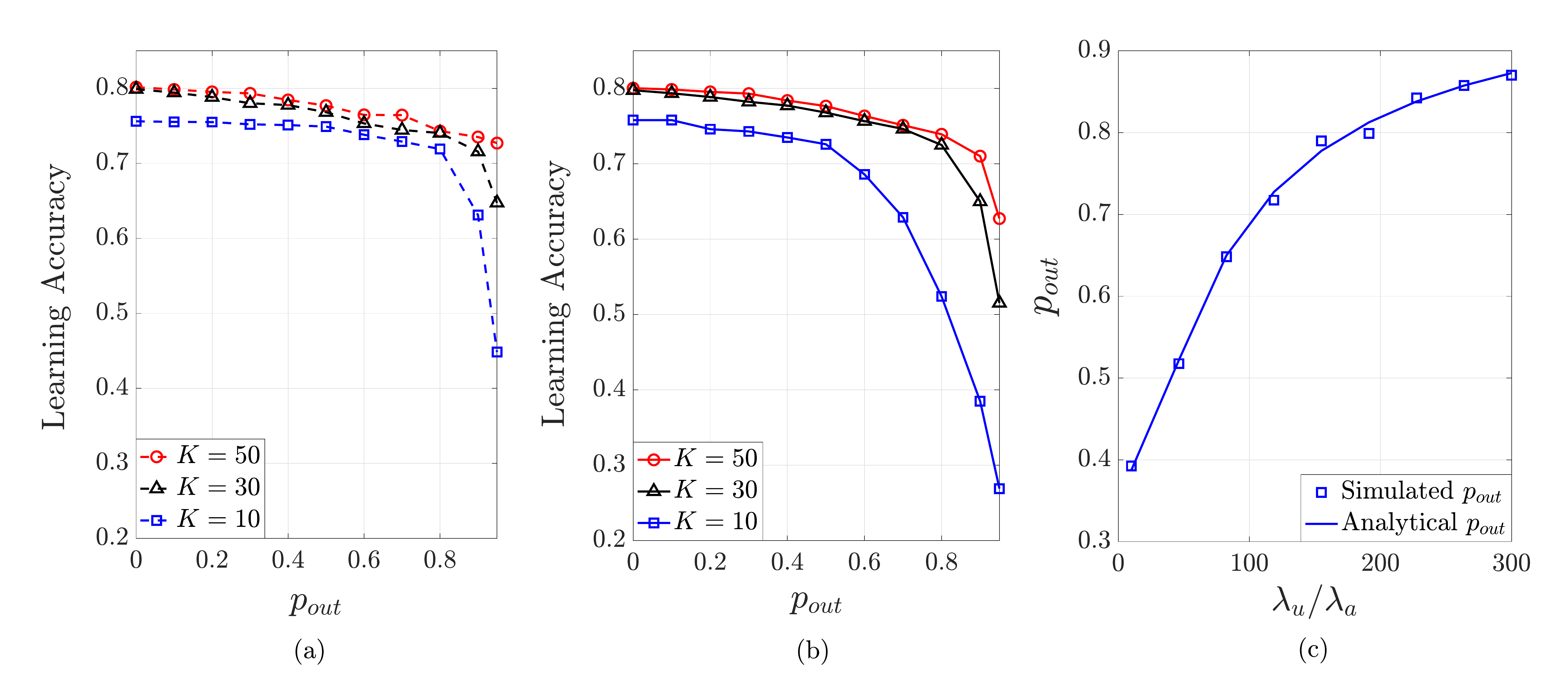}
		\centering
		\caption{The numerical results of the proposed intermittent FL: (a)  Uplink outage probability versus learning accuracy for different numbers of UAVs with i.i.d. datasets; (b) Uplink outage probability versus learning accuracy for different numbers of UAVs with non-i.i.d. datasets; (c) Uplink outage probability versus the ratio of the UAV density to the AP density.}
		\label{Fig:AccUplinkOutProb}
		\vspace{-0.2in}
\end{figure} 

The numerical results of the proposed intermittent FL are shown in Fig.~\ref{Fig:AccUplinkOutProb}. Specifically, Figs.~\ref{Fig:AccUplinkOutProb} (a) and (b) show how the learning accuracy varies with $p_{out}$ for the different numbers of the UAVs with i.i.d. and non-i.i.d. datasets, respectively. As can be seen in Fig.~\ref{Fig:AccUplinkOutProb}(a), the learning accuracy reduces as $p_{out}$ increases, yet it improves as the average number of the UAVs in the network increases. As $p_{out}$ increases, the uplink communication outage is more likely to happen and thus the edge server more likely does the global model aggression with less local learning outcomes, which essentially slows down the convergence process of FL. As a result, the global model vector $w_t$ is less likely to converge to a stable vector within $T$ training rounds. This is why the learning accuracy reduces as $p_{out}$ increases. When more UAVs with i.i.d. datasets join FL, more local training outcomes with a similar statistical distribution are likely to be aggregated at the edge, which improves the convergence rate of FL. Therefore,  increasing the average number of the UAVs with i.i.d. datasets in the network helps to improve the learning accuracy because it mitigates the negative impact of the uplink communication outage on the convergence process of FL. Moreover,  Fig.~\ref{Fig:AccUplinkOutProb}(b) reveals a phenomenon different from Fig.~\ref{Fig:AccUplinkOutProb}(a), that is, the learning accuracy is more sensitive to the average number of the UAVs with non-i.i.d. datasets in the network. For example, when $p_{out}=0.6$, increasing $K$ from $10$ to $50$ significantly improves the learning accuracy, whereas we cannot observe this in Fig.~\ref{Fig:AccUplinkOutProb}(a). This phenomenon stems from the fact that \textit{every} local learning outcome is crucial to the global model aggregation at the edge server in that the local learning outcomes obtained from non-i.i.d. datasets may have a very distinct statistical distribution. Hence, the negative impact of non-i.i.d. datasets on the convergence process of FL can be effectively mitigated by increasing the average number of the UAVs. Fig.~\ref{Fig:AccUplinkOutProb}(c) presents how $p_{out}$ increases along with $\lambda_u/\lambda_a$. When $\lambda_u/\lambda_a$ increases, more UAVs are in the network and more uplink interference is generated, thereby making uplink communication outage more likely occur. The results in this figure provide a fundamental relationship between $p_{out}$ and $\lambda_u/\lambda_a$ and they can be used together with the results in Figs.~\ref{Fig:AccUplinkOutProb}(a) and (c) to provide some insight into how to deploy APs and UAVs in order to achieve a desired learning accuracy for a given uplink outage probability. For instance, we can deploy APs and UAVs with a ratio of $\lambda_u/\lambda_a\approx 100$ in order to achieve a learning accuracy about $0.75$ by deploying the average number of the UAVs with non-i.i.d. datasets greater than $30$.

%In the following section, we will demonstrate the numerical results and discuss how deployments of the UAVs and APs affect such an FL algorithm that models the outage of data delivery between the UAV and its associating AP proposed in Sec. \ref{SubSec:FedLearnPreSim}.

\begin{figure}[!t]
	\centering
	\includegraphics[width=3.5in, height=1.85in]{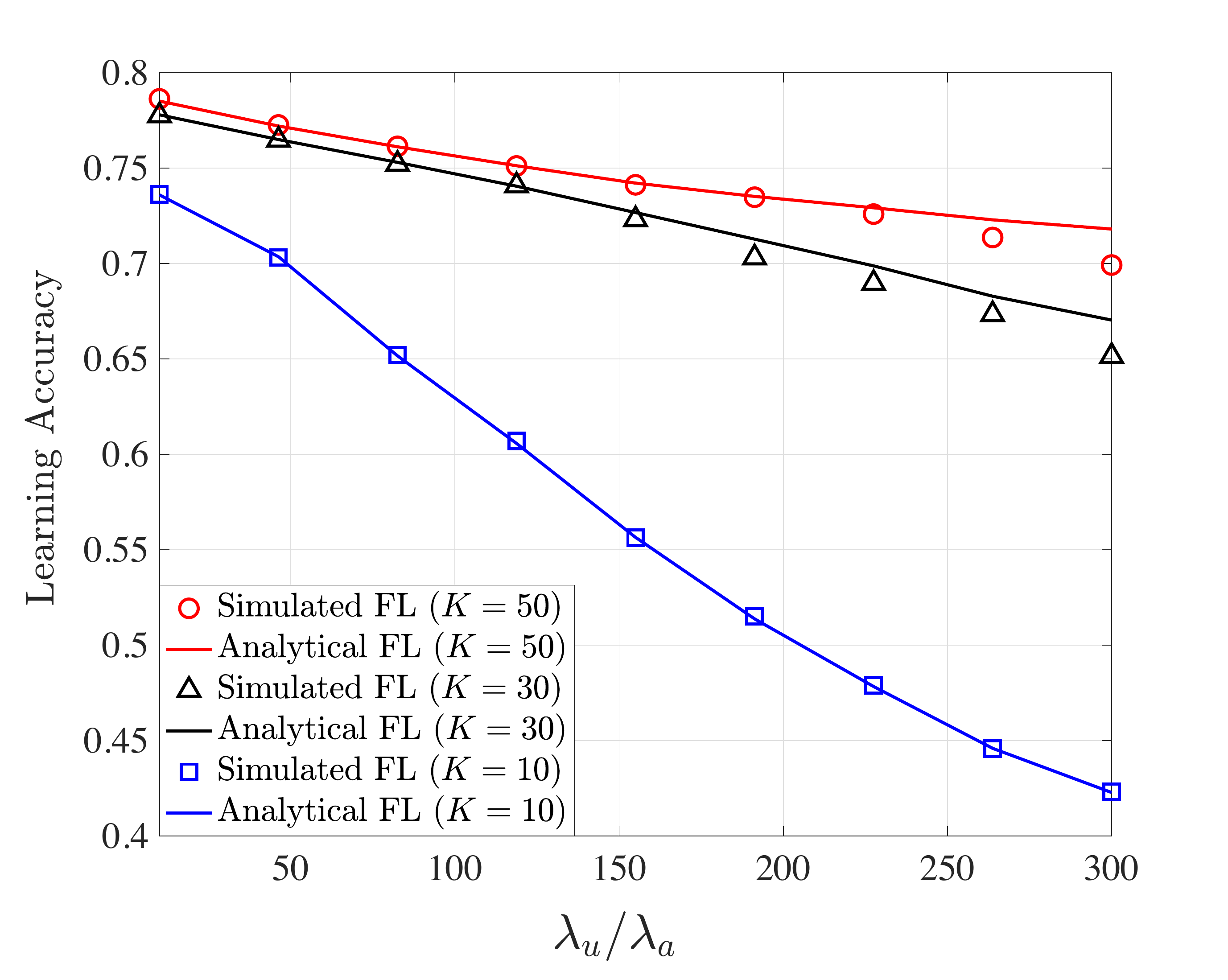}
	\vspace{-0.25in}
	\caption{Simulation results of the learning accuracy of the proposed FL model for the case of non-i.i.d. datasets at the UAVs.}
	\label{Fig:AccLambda}
	\vspace{-0.15in}
\end{figure}

\subsection{Numerical Results of the Learning Accuracy vs. $\lambda_u/\lambda_a$}\label{SubSec:SimulationOutProb}
	
Fig.~\ref{Fig:AccLambda} shows how the learning accuracy varies with $\lambda_u/\lambda_a$ when all the UAVs have non-i.i.d. datasets. Two cases of simulated FL and analytical FL are presented in the figure. The results of the simulated FL are completely obtained by running numerical simulations, yet the results of the analytical FL are obtained by first calculating $p_{out}$ based on the expression in~\eqref{Eqn:UplinkOutProb} for different values of $\lambda_u/\lambda_a$ and then using Fig.~\ref{Fig:AccUplinkOutProb}(b) to find the values of learning accuracy corresponding to the calculated values of $p_{out}$. As can be seen in the figure, the results of analytical FL almost coincide with their corresponding results of simulated FL. This reveals the correctness and accuracy of the uplink outage probability found in~\eqref{Eqn:UplinkOutProb}. Furthermore, the results in Fig.~\ref{Fig:AccLambda} demonstrate the fact that the performance of FL over a UAV network can indeed be impacted by how densely the APs and UAVs are deployed in the network. This is an important and interesting finding worth further investigation. 
	
\section{Conclusion}\label{Sec:Conclusion}
In the literature, the studies of FL over wireless network were mainly conducted based on a unrealistic assumption, i.e., no communication outage between clients and a server when conducting FL. Such studies cannot practically reflect the accurate performance of FL over wireless networks. To understand how communication outage impacts FL, this paper proposed an intermittent FL model that is able to characterize uplink communication outages in a cellular-connected UAV network. A tractable approach to analyzing the uplink outage probability was proposed and the uplink outage probability of a UAV was explicitly derived in a neat form. We found that the performance of FL over a UAV network can be significantly impacted by the uplink outage probability that depends on how the APs and UAVs are deployed in the network. Numerical results not only validate the accuracy of the analyses of the uplink outage probability, but also support the finding on how the uplink outage probability degrades the the performance of FL over unreliable wireless networks.  
	
\appendix 
\subsection{Proof of Theorem \ref{Thm:CDFAssoPower}}\label{App:ProofAssoPower}
According to the definition of $R_{i\star}$(r), $R_{i\star}(r)$ for a given $H_i$ can be further written as
\begin{align*}
R_{i\star}(r) &\stackrel{(a)}{=} \mathbb{P}\left[\max_{j:A_j\in\Phi_a} \left\{\frac{ L_{ij}}{[\|X_i-A_j\|^2+H_i^2]^{\frac{\alpha}{2}}}\right\}\leq r\right]\\
&\stackrel{(b)}{=} \mathbb{E}\left\{\prod_{j:A_j\in\Phi_a}\mathbb{P}\left[\frac{L_{ij}}{[\|A_j\|^2+H_i^2]^{\frac{\alpha}{2}}}\leq r\right]\right\}\\
&\stackrel{(c)}{=}\exp\left(-2\pi\lambda_a\int^\infty_0\mathbb{P}\left[\frac{L}{\left[x^2+H^2\right]^{\frac{\alpha}{2}}}> r\right]x\mathrm{d}x\right),
\end{align*}
where (a) is obtained based on the UAV association scheme in~\eqref{Eqn:AssoScheme}, (b) is obtained by considering $X_j$ as the origin and the fact that all $L_{ij}/(\|A_j\|^2+H_j^2)^{\alpha/2}$~'s are independent, and (c) is obtained by applying probability generating functional (PGFL) of an HPPP to $\Phi_a$. The subscript $ij$ of $L_{ij}$ is dropped for simplifying notation. Replacing $x^2$ with $y$ and letting $\vartheta_y = \tan^{-1}(H_j/\sqrt{y})$ yield the following:
\begin{align*}
&\mathbb{P}\left[\frac{L}{\left[y+H_i^2\right]^{\frac{\alpha}{2}}}> r\right] = \rho(\vartheta_i(y))\mathbb{P}\left[\frac{1}{(y+H_i^2)^\frac{\alpha}{2}}>r\right]\\
&+[1-\rho(\vartheta_i(y))]\mathbb{P}\left[\frac{\ell}{(y+H_i^2)^\frac{\alpha}{2}}>r\right]= \rho(\vartheta_i(y))\times\\
&\mathbb{P}\bigg[y<r^{-\frac{2}{\alpha}}-H_i^2\bigg]+[1-\rho(\vartheta_i(y))]\mathbb{P}\left[y<\left(\frac{\ell}{r}\right)^\frac{2}{\alpha}-H_i^2\right].
\end{align*}
Thus, $\Upsilon_i(r)=\int_0^{\infty} \mathbb{P}\left[L(y+H_i^2)^{-\frac{\alpha}{2}}> r\right]\dif y$ and $R_{i\star}(r)$ are obtained accordingly.
	
\subsection{Proof of Theorem~\ref{Thm:LapTranUplinkInter}}\label{App:ProofLapTranUplinkInter}
According to the definition of $I_{\star}$ and $\|U_k\|^2=\|X_k\|^2+\|H_k\|^2$, the Laplace transform of $I_{\star}$ can be further written as
\begin{align*}
&\mathbb{E}\left[e^{-sI_{\star}}\right] = \mathbb{E}\left[\prod_{i:U_i\in\widetilde{\Phi}_u\backslash U_\star}\exp\left(-\frac{sG_{k\star}L_{k\star}}{(\|X_k\|^2+H_k^2)^{\frac{\alpha}{2}}}\right)\right]\\
&\stackrel{(a)}{=}\exp\left(-\pi\lambda_a\int^\infty_0\left\{1-\mathbb{E}\left[e^{-\frac{sGL}{(y+H^2)^{\alpha/2}}}\right]\right\}\mathrm{d}y\right)\\
&\stackrel{(b)}{=}\exp\left(-\pi\lambda_a\int^\infty_0\left\{1-\mathbb{E}\left[\mathcal{L}_G\left(\frac{sL}{(y+H^2)^{\frac{\alpha}{2}}}\right)\right]\right\}\mathrm{d}y\right),
\end{align*}
where $(a)$ is obtained by applying the PGFL of an HPPP to the projections of $\widetilde{\Phi}_u$ and using $G$ instead of $G_{k\star}$. Since $\vartheta(y) =\tan^{-1}(H/\sqrt{y})$, we thus have
\begin{align*}
&\mathbb{E}\left[\mathcal{L}_G\left(\frac{sL}{(y+H^2)^{\frac{\alpha}{2}}}\right)\right]= \mathbb{E}\left[\mathcal{L}_G\left(sLy^{-\frac{\alpha}{2}}\sec^{\alpha}(\vartheta(y))\right)\right]=\\
&\rho(\vartheta_i(y))\mathbb{E}\left[\mathcal{L}_G\left(sy^{-\frac{\alpha}{2}}\sec^{\alpha}(\vartheta(y))\right)\right]+ [1-\rho(\vartheta_i(y))]\\
&\times\mathbb{E}\left[\mathcal{L}_G\left(sy^{-\frac{\alpha}{2}}\ell\sec^{\alpha}(\vartheta(y))\right)\right].
\end{align*}

% reference section
\bibliographystyle{IEEEtran}
\bibliography{IEEEabrv,RefFedLearnUAV}
	
\end{document}